\documentclass[10pt, a4paper]{article}
\usepackage{lrec}
\usepackage{subfig}
\usepackage{booktabs}
\usepackage{multirow}
\usepackage{graphicx}
\usepackage{tabularx}
\usepackage[utf8]{inputenc}
\usepackage[english]{babel}
\usepackage{hyperref}
\usepackage{times}
\usepackage{latexsym}
\usepackage{times}
\usepackage{url}
\usepackage{dblfloatfix}
\usepackage[colorinlistoftodos]{todonotes} 
\usepackage{multicol}
\usepackage{dblfloatfix}

\title{Named Entities in Medical Case Reports: Corpus and Experiments}

\name{Sarah Schulz\textsuperscript{1}, Jurica \v{S}eva\textsuperscript{1}, Samuel Rodriguez\textsuperscript{1}, Malte Ostendorff\textsuperscript{2}, Georg Rehm\textsuperscript{2}}

\address{\textsuperscript{1}Ada Health GmbH, Karl-Liebknecht-Str. 1, 10178 Berlin, Germany\\
        \{first.lastname\}@ada.com\\
        \textsuperscript{2}DFKI GmbH, Alt-Moabit 91c, 10559 Berlin, Germany\\
       \{first.lastname\}@dfki.de\\
}

\abstract{
We present a new corpus comprising annotations of medical entities in case reports, originating from PubMed Central's open access library. In the case reports, we annotate cases, conditions, findings, factors and negation modifiers. Moreover, where applicable, we annotate relations between these entities. As such, this is the first corpus of this kind made available to the scientific community in English. It enables the initial investigation of automatic information extraction from case reports through tasks like Named Entity Recognition, Relation Extraction and (sentence/paragraph) relevance detection. 
Additionally, we present four strong baseline systems for the detection of medical entities made available through the annotated dataset. 
\\ \newline \Keywords{Named Entity Recognition, Case Reports, Corpus} } 

\begin{document}


\maketitleabstract

\section{Introduction}

The automatic processing of medical texts and documents plays an increasingly important role in the recent development of the digital health area. 
To enable dedicated Natural Language Processing (NLP) that is highly accurate with respect to medically relevant categories, manually annotated data from this domain is needed. 
One category of high interest and relevance are medical entities. Only very few annotated corpora in the medical domain exist. 
Many of them focus on the relation between chemicals and diseases or proteins and diseases, such as the BC5CDR corpus \cite{BC5}, the Comparative Toxicogenomics Database\footnote{\url{http://ctdbase.org}} ~\cite{davis2019comparative}, the FSU PRotein GEne corpus\footnote{\url{https://julielab.de/Resources/FSU_PRGE.html}} ~\cite{fsuprge} or the ADE (adverse drug effect) corpus ~\cite{gurulingappa2012}.
The NCBI Disease Corpus \cite{Dogan:2014:NDC:2772763.2772800} contains condition mention annotations along with annotations of symptoms. 
Several new corpora of annotated case reports were made available recently. \newcite{grouin-etal-2019-clinical} presented a corpus with medical entity annotations of clinical cases written in French, \newcite{copdPhenotype} presented a corpus focusing on phenotypic information for chronic obstructive pulmonary disease while ~\newcite{10.1093/database/bay143} presented a corpus focusing on  identifying main finding sentences in case reports.

\begin{table*}[!hbt]
\resizebox{\textwidth}{!}{%
\begin{tabular}{llll}
\hline
Corpus &
  Annotated entities &
  Relationships &
  \# documents \\ \hline
BC5CDR &
  \begin{tabular}[c]{@{}l@{}}chemicals (4,409), diseases (5,818)\end{tabular} &
  chemical-disease (3116) &
  1,500 PubMed articles \\ 
FSU PRotein GEne &
  \begin{tabular}[c]{@{}l@{}}protein, protein\_familiy\_or\_group, protein\_complex, protein\_variant, protein\_enum\end{tabular} &
  - &
  3,308 PubMed articles \\ 
ADE &
  \begin{tabular}[c]{@{}l@{}}drugs (5,063), adverse effect (5,776), dosage (231)\end{tabular} &
  \begin{tabular}[c]{@{}l@{}}drug-adverse effect (6821), drug-dosage (279)\end{tabular} &
  3,000 PubMed articles \\ 
NCBI Disease Corpus &
  diseases (6,892) &
  - &
  793 PubMed abstracts \\ 
Grouin et al. &
  31 targets &
  - &
  717 clinical cases \\
COPD &
  16 COPD phenotype targets &
  - &
  30 full texts \\ 
Smalheiser et al. &
  sentenc(es) with main finding(s) &
  - &
  416 full texts \\ 
Our work &
  case (69), condition (347), factor (363), finding (3,248), modifier (336) &
  - &
  53 full texts \\ \hline
\end{tabular}
}
\caption{Summary overview of relevant and comparable corpora.}
\label{corporaOverview}
\end{table*}

The corpus most comparable to ours is the French corpus of clinical case reports by ~\newcite{grouin-etal-2019-clinical}. 
Their annotations are based on UMLS semantic types. Even though there is an overlap in annotated entities, semantic classes are not the same. Lab results are subsumed under findings in our corpus and are not annotated as their own class. 
Factors extend beyond gender and age and describe any kind of risk factor that contributes to a higher probability of having a certain disease. Our corpus includes additional entity types. We annotate \textit{conditions}, \textit{findings} (including medical findings such as blood values), \textit{factors}, 
and also \textit{modifiers} which indicate the negation of other entities as well as \textit{case entities}, i.\,e., entities specific to one case report. 
An overview is available in Table ~\ref{corporaOverview}.

\section{A Corpus of Medical Case Reports with Medical Entity Annotation}

\subsection{Annotation tasks}

Case reports are standardized in the CARE guidelines \cite{Rison2013}. 
They represent a detailed description of the symptoms, signs, diagnosis, treatment, and follow-up of an individual patient. 
We focus on documents freely available through PubMed Central\footnote{https://www.ncbi.nlm.nih.gov/pmc/} (PMC). 
The presentation of the patient's case can usually be found in a dedicated section or the abstract. 
We perform a manual annotation of all mentions of case entities, conditions, findings, factors and modifiers. 
The scope of our manual annotation is limited to the presentation of a patient's signs and symptoms. In addition, we annotate the title of the case report.

\subsection{Annotation Guidelines}
\label{ann_guid}

\begin{figure}
\centering
 \subfloat[Factor and case annotation\label{factorcase}]{%
  \includegraphics[width=0.45\textwidth]{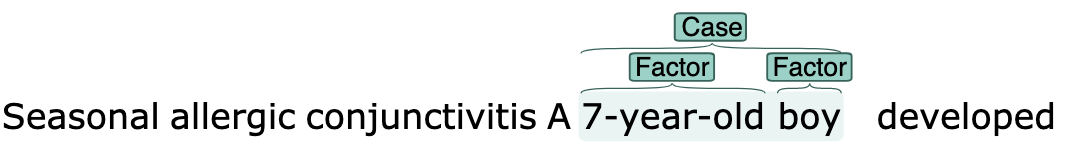}
 }
 \hfill
 \subfloat[Factor and condition annotation\label{faccon}]{%
  \includegraphics[width=0.5\textwidth]{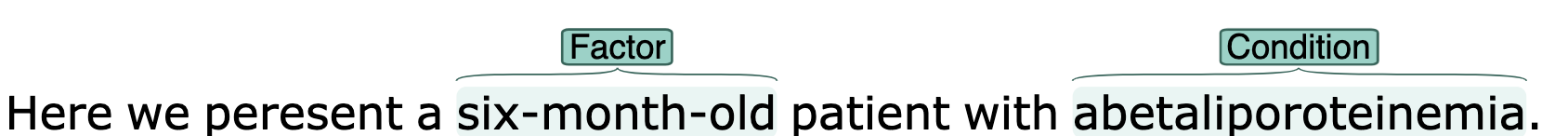}
 }
 \hfill
 \subfloat[Partially nested finding annotations\label{partially_nested}]{%
  \includegraphics[width=0.5\textwidth]{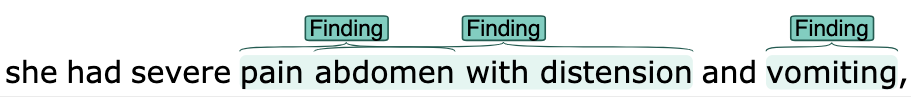}
 }
\caption{Annotated entities (WebAnno)}
\label{anno}
\end{figure}

We annotate the following entities:
\begin{itemize}
\itemsep0em 
\item \textbf{case entity} marks the mention of a patient. A case report can contain more than one case description. Therefore, all the findings, factors and conditions related to one patient are linked to the respective case entity. Within the text, this entity is often represented by the first mention of the patient and overlaps with the factor annotations which can, e.\,g., mark sex and age (cf.~Figure~\ref{anno}).
\item \textbf{condition} marks a medical disease such as \textit{pneumothorax} or \textit{dislocation of the shoulder}.
\item \textbf{factor} marks a feature of a patient which might influence the probability for a specific diagnosis. It can be immutable (e.\,g., \textit{sex} and \textit{age}), describe a specific medical history (e.\,g., \textit{diabetes mellitus}) or a behaviour (e.\,g., \textit{smoking}).
\item \textbf{finding} marks a sign or symptom a patient shows. This can be visible (e.\,g., \textit{rash}), described by a patient (e.\,g., \textit{headache}) or measurable (e.\,g., \textit{decreased blood glucose level}).
\item \textbf{negation modifier} explicitly negate the presence of a certain finding usually setting the case apart from common cases. 
\end{itemize}


We also annotate relations between these entities, where applicable. Since we work on case descriptions, the anchor point of these relations is the case that is described. The following relations are annotated:

\begin{itemize}
\itemsep0em 
\item \textbf{has} relations exist between a case entity and factor, finding or condition entities.
\item \textbf{modifies} relations exist between negation modifiers and findings.
\item \textbf{causes} relations exist between conditions and findings.
\end{itemize}

\begin{figure}
\centering
     \subfloat[Annotation of relation \textit{has}\label{has}]{%
       \includegraphics[width=0.45\textwidth]{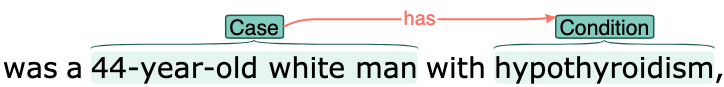}
     }
     \hfill
      \subfloat[Annotation of relation \textit{modifies}\label{modifies}]{%
       \includegraphics[width=0.45\textwidth]{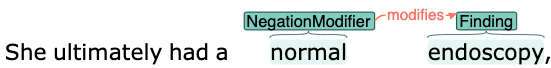}
     }
     \hfill
     \subfloat[Annotation of relation \textit{causes}\label{causes}]{%
       \includegraphics[width=0.5\textwidth]{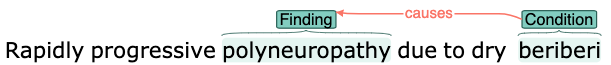}
     }
\caption{Annotated relations between entities (WebAnno)}
\label{relations}
\end{figure}

Example annotations are shown in Figure~\ref{relations}.

\subsection{Annotators}

We asked medical doctors experienced in extracting knowledge related to medical entities from texts to annotate the entities described above. 
Initially, we asked four annotators to test our guidelines on two texts. 
Subsequently, identified issues were discussed and resolved. 
Following this pilot annotation phase, we asked two different annotators to annotate two case reports according to our guidelines. 
The same annotators annotated an overall collection of 53 case reports. 

Inter-annotator agreement is calculated based on two case reports. We reach a Cohen's kappa \cite{Cohen1960} of 0.68. Disagreements mainly appear for findings that are rather unspecific such as \textit{She no longer eats out with friends} which can be seen as a finding referring to ``avoidance behaviour''. 

\subsection{Annotation Tools and Format}

The annotation was performed using WebAnno\footnote{\url{https://webanno.github.io/webanno/}, 08/04/2019.} \cite{eckart-de-castilho-etal-2016-web}, a web-based tool for linguistic annotation. The annotators could choose between a pre-annotated version or a blank version of each text. The pre-annotated versions contained suggested entity spans based on string matches from lists of conditions and findings synonym lists. Their quality varied widely throughout the corpus. The blank version was preferred by the annotators.
We distribute\footnote{The corpus can be dowloaded here: \url{https://github.com/adahealth/medical_case_report_corpus}.} the corpus in 
BioC JSON format\footnote{\url{http://bioc.sourceforge.net}}.
BioC was chosen as it allows us to capture the complexities of the annotations in the biomedical domain. 
It represented each documents properties ranging from full text, individual passages/sentences along with captured annotations and relationships in an organized manner. 
BioC is based on character offsets of annotations and allows the stacking of different layers. 

\subsection{Corpus Overview}


The corpus consists of 53 documents, which contain an average number of 156.1 sentences per document, each with 19.55 tokens on average. The corpus comprises 8,275 sentences and 167,739 words in total.
\footnote{Sentence splitting and tokenization are performed using ScispaCy (\url{https://allenai.github.io/scispacy/}) and its \textit{en\_core\_sci\_md} model} However, as mentioned above, only case presentation sections, headings and abstracts are annotated. The numbers of annotated entities are summarized in Table~\ref{corpusstat}.

\begin{table}[!htb]
\resizebox{\columnwidth}{!}{%
\begin{tabular}{lrrrr} \toprule
Type & Number & Max & Min & Mean\\ \midrule
Documents & 53 & -- & -- & -- \\
Sentences & 8,275 & 827 & 44 & 156.1 \\ 
Words & 167,739& 16,309& 1,260 &3164.9\\ 
Annotated sentences & 1063 & 228 & 1 & 19.55 \\ 
\midrule
case & 69 & 5 & 1 & 3.1\\
condition & 347 &9&1 &2.0\\
factor & 363&16&1&2.5\\
finding & 3,248& 25& 1& 2.6\\
modifier & 336 &18 &1&1.4\\
\midrule
total annotations & 4,363 & -- & -- & -- \\
discontinuous & 1,055 & -- & -- & -- \\
multi-label & 1,535 & -- & -- & -- \\
discontinuous and multi-label & 541 & -- & -- & -- \\
nested & 603 & -- & -- & -- \\
fully nested & 584 & -- & -- & -- \\
partially nested & 19 & -- & -- & -- \\ \bottomrule
\end{tabular}
}
\caption{Corpus statistics}
\label{corpusstat}
\end{table}

Findings are the most frequently annotated type of entity. 
This makes sense given that findings paint a clinical picture of the patient's condition. 
The number of tokens per entity ranges from one token for all types to 5 tokens for cases (average length 3.1), nine tokens for conditions (average length 2.0), 16 tokens for factors (average length 2.5), 25 tokens for findings (average length 2.6) and 18 tokens for modifiers (average length 1.4) (cf.~Table~\ref{corpusstat}). Examples of rather long entities are given in Table~\ref{entityex}. 

\begin{table}[h]
  \begin{small}
  \begin{tabular}{lp{6cm}} \toprule
  Type & Example\\ \midrule
  case & 42-year-old poorly prepared mountaineer\\
  condition & Salter–Harris type II epiphysiolysis at the proximal left humerus\\
  factor & 5 ml of paracetamol was given to child every 4 hours for the past 6 days\\
  finding & nests and sheets of cells with moderate-to-abundant cytoplasm, eccentrically placed nuclei surrounded by dense pink homogeneous material\\
  modifier & height and weight were at the 25th - 50th  percentile and the 50th - 75th  percentile\\ \bottomrule
  \end{tabular}
  \end{small}  
  \caption{Examples of long entities per type}
  \label{entityex}
\end{table}

\begin{figure}[h]
\centering
\includegraphics[width=0.45\textwidth]{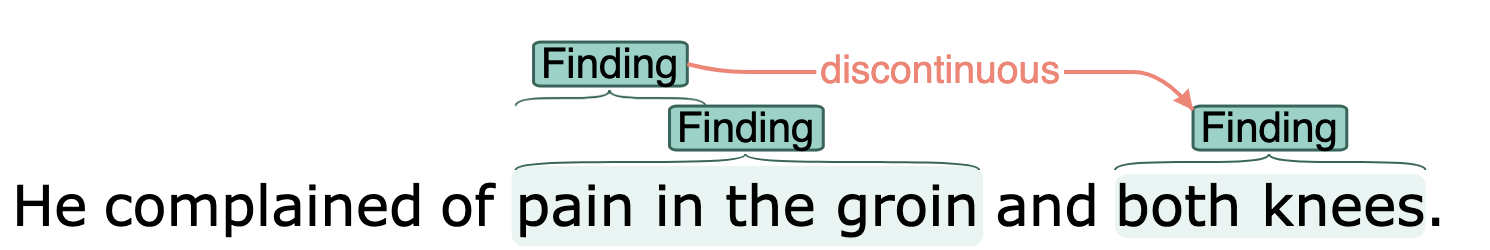}
\caption{Discontinuous finding annotation\label{discontinuous}}
\end{figure}

Entities can appear in a discontinuous way. We model this as a relation between two spans which we call ``discontinuous'' (cf.~Figure~\ref{discontinuous}). Especially findings often appear as discontinuous entities, we found 543 discontinuous finding relations. The numbers for conditions and factors are lower with seven and two, respectively. Entities can also be nested within one another. This happens either when the span of one annotation is completely embedded in the span of another annotation (fully-nested; cf.~Figure~\ref{factorcase}), or when there is a partial overlapping between the spans of two different entities (partially-nested; cf.~Figure~\ref{partially_nested}). There is a high number of inter-sentential relations in the corpus (cf.~Table~\ref{relationstats}). This can be explained by the fact that the case entity occurs early in each document; furthermore, it is related to finding and factor annotations that are distributed across different sentences.

\begin{table}[!htb]
\resizebox{\columnwidth}{!}{%
\begin{tabular}{lrrrrrr} \toprule
Type of Relation & \multicolumn{2}{c}{Intra-sentential} &\multicolumn{2}{c}{Inter-sentential} & \multicolumn{2}{c}{Total}\\ \midrule
case has condition & 28 & 18.1\% & 127 & 81.9\% & 155 & 4.0\%\\
case has finding & 169 & 7.2\% & 2180 & 92.8\% & 2349 & 61.0\%\\
case has factor & 153 & 52.9\% & 136 &  47.1\% & 289 & 7.5\%\\
modifier modifies finding & 994 & 98.5\% & 15 & 1.5\% & 1009 & 26.2\%\\
condition causes finding &44 & 3.6\% & 3 & 6.4\% & 47 & 1.2\%\\ \bottomrule
\end{tabular}
}
\caption{Annotated relations between entities. Relations appear within a sentence (intra-sentential) or across sentences (inter-sentential)}
\label{relationstats}
\end{table}

The most frequently annotated relation in our corpus is the \textit{has}-relation between a case entity and the findings related to that case. This correlates with the high number of finding entities. The relations contained in our corpus are summarized in Table~\ref{relationstats}.


\section{Baseline systems for Named Entity Recognition in medical case reports}

We evaluate the corpus using Named Entity Recognition (NER), i.\,e., the task of finding mentions of concepts of interest in unstructured text. 
We focus on detecting cases, conditions, factors, findings and modifiers in case reports (cf.~Section~\ref{ann_guid}). We approach this as a sequence labeling problem. Four systems were developed to offer comparable robust baselines. 

The original documents are pre-processed (sentence splitting and tokenization with ScispaCy\footnote{\url{https://github.com/allenai/scispacy/releases/tag/v0.2.2}~\cite{neumann-etal-2019-scispacy} and the \textit{en\_core\_sci\_md} model.}). 
We do not perform stop word removal or lower-casing of the tokens. 
The BIO labeling scheme is used to capture the order of tokens belonging to the same entity type and enable span-level detection of entities. 
Detection of nested and/or discontinuous entities is not supported.  
The annotated corpus is randomized and split in five folds using scikit-learn \cite{sklearn_api}. 
Each fold has a train, test and dev split with the test split defined as .15\% of the train split. 
This ensures comparability between the presented systems. 

\begin{table*}[!ht]
\centering
\begin{tabular}{|l|r|r|r|l|l|l|l|l|l|r|r|r|}
\hline
\multirow{2}{*}{} & \multicolumn{3}{c|}{CRF} & \multicolumn{3}{c|}{BiLSTM CRF} & \multicolumn{3}{c|}{MTL} & \multicolumn{3}{c|}{BioBERT} \\ \cline{2-13} 
 & \multicolumn{1}{c|}{P} & \multicolumn{1}{c|}{R} & \multicolumn{1}{c|}{F1} & \multicolumn{1}{c|}{P} & \multicolumn{1}{c|}{R} & \multicolumn{1}{c|}{F1} & \multicolumn{1}{c|}{P} & \multicolumn{1}{c|}{R} & \multicolumn{1}{c|}{F1} & \multicolumn{1}{c|}{P} & \multicolumn{1}{c|}{R} & \multicolumn{1}{c|}{F1} \\ \hline
case & 0.59 & 0.76 & \textbf{0.66} & 0.40 & 0.22 & 0.28 & 0.55 & 0.38 & 0.44 & 0.43 & 0.64 & 0.51 \\ \hline
condition & 0.45 & 0.18 & 0.26 & 0.00 & 0.00 & 0.00 & 0.62 & 0.62 & \textbf{0.62} & 0.33 & 0.37 & 0.34 \\ \hline
factor & 0.40 & 0.05 & 0.09 & 0.23 & 0.04 & 0.06 & 0.6 & 0.53 & \textbf{0.56} & 0.17 & 0.10 & 0.12 \\ \hline
finding & 0.50 & 0.33 & 0.40 & 0.39 & 0.26 & 0.31 & 0.62 & 0.61 & \textbf{0.61} & 0.41 & 0.53 & 0.46 \\ \hline
modifier & 0.74 & 0.32 & 0.45 & 0.60 & 0.42 & 0.47 & 0.66 & 0.63 & \textbf{0.65} & 0.51 & 0.52 & 0.50 \\ \hline
micro avg. & 0.52 & 0.31 & 0.39 & 0.41 & 0.23 & 0.30 & 0.52 & 0.44 & \textbf{0.47} & 0.39 & 0.49 & 0.43 \\ \hline
macro avg. & 0.51 & 0.31 & 0.38 & 0.37 & 0.23 & 0.28 & \multicolumn{1}{c|}{0.61} & \multicolumn{1}{c|}{0.58} & \multicolumn{1}{c|}{\textbf{0.59}} & 0.40 & 0.49 & 0.44 \\ \hline

\end{tabular}
\caption{\label{font-table} Span-level precision (P), recall (R) and F1-scores (F1) on four distinct baseline NER systems. All scores are computed as average over five-fold cross validation.}
\label{results}
\end{table*}

\subsection{Conditional Random Fields}

Conditional Random Fields (CRF) \cite{Lafferty:2001:CRF:645530.655813} are a standard approach when dealing with sequential data in the context of sequence labeling. We use a combination of linguistic and semantic features\footnote{A list of features will be published with the corpus to guarantee reproducibility of the results.}, with a context window of size five, to describe each of the tokens and the dependencies between them. Hyper-parameter optimization is performed using randomized search and cross validation. Span-based F1 score is used as the optimization metric. 

\subsection{BiLSTM-CRF}
\label{ss:bilstm}

Prior to the emergence of deep neural language models, BiLSTM-CRF models~\cite{huang2015bidirectional} had achieved state-of-the-art results for the task of sequence labeling. We use a BiLSTM-CRF model with both word-level and character-level input. 
BioWordVec\footnote{\url{https://ftp.ncbi.nlm.nih.gov/pub/lu/Suppl/BioSentVec/BioWordVec_PubMed_MIMICIII_d200.bin}} \cite{chen2018biosentvec} pre-trained word embeddings are used in the embedding layer for the input representation. 
A bidirectional LSTM layer is applied to a multiplication of the two input representations. Finally, a CRF layer is applied to predict the sequence of labels. 
Dropout and L1/L2 regularization is used where applicable. He (uniform) initialization~\cite{He:2015:DDR:2919332.2919814} is used to initialize the kernels of the individual layers. As the loss metric, CRF-based loss is used, while optimizing the model based on the CRF Viterbi accuracy. Additionally, span-based F1 score is used to serialize the best performing model. 
We train for a maximum of 100 epochs, or until an early stopping criterion is reached (no change in validation loss value grater than 0.01 for ten consecutive epochs). Furthermore, Adam \cite{Kingma2014AdamAM} is used as the optimizer. The learning rate is reduced by a factor of 0.3 in case no significant increase of the optimization metric is achieved in three consecutive epochs. 

\subsection{Multi-Task Learning}

Multi-Task Learning (MTL) \cite{ruder2017overview} has become popular with the progress in deep learning. This model family is characterized by simultaneous optimization of multiple loss functions and transfer of knowledge achieved this way. The knowledge is transferred through the use of one or multiple shared layers. Through finding supporting patterns in related tasks, MTL provides better generalization on unseen cases and the main tasks we are trying to solve. 

We rely on the model presented by \newcite{bekoulis2018joint} and reuse the implementation provided by the authors.\footnote{\url{https://github.com/bekou/multihead\_joint\_entity\_relation\_extraction}} The model jointly trains two objectives supported by the dataset: the main task of NER and a supporting task of Relation Extraction (RE). Two separate models are developed for each of the tasks. The NER task is solved with the help of a BiLSTM-CRF model, similar to the one presented in Section~\ref{ss:bilstm} 
The RE task is solved by using a multi-head selection approach, where each token can have none or more relationships to in-sentence tokens. Additionally, this model also leverages the output of the NER branch model (the CRF prediction) to learn label embeddings. Shared layers consist of a concatenation of word and character embeddings followed by two bidirectional LSTM layers. We keep most of the parameters suggested by the authors and change (1) the number of training epochs to 100 to allow the comparison to other deep learning approaches in this work, (2) use label embeddings of size 64, (3) allow gradient clipping and (4) use $d=0.8$ as the pre-trained word embedding dropout and $d=0.5$ for all other dropouts. $\eta=1^{-3}$ is used as the learning rate with the Adam optimizer and \textit{tanh} activation functions across layers. Although it is possible to use adversarial training \cite{bekoulis2018adversarial}, we omit from using it. We also omit the publication of results for the task of RE as we consider it to be a supporting task and no other competing approaches have been developed. 

\subsection{BioBERT} 

Deep neural language models have recently evolved to a successful method for representing text. In particular, Bidirectional Encoder Representations from Transformers (BERT) outperformed previous state-of-the-art methods by a large margin on various NLP tasks \cite{Devlin2018}. For our experiments, we use BioBERT, an adaptation of BERT for the biomedical domain, pre-trained on PubMed abstracts and PMC full-text articles \cite{Lee2019}. The BERT architecture\footnote{We use the PyTorch version by HuggingFace \cite{Wolf2019}.} for deriving text representations uses 12 hidden layers, consisting of 768 units each. For NER, token level BIO-tag probabilities are computed with a single output layer based on the representations from the last layer of BERT. We fine-tune the model on the entity recognition task during four training epochs with batch size $b=32$, dropout probability $d=0.1$ and learning rate $\eta=2^{-5}$. These hyper-parameters are proposed by \newcite{Devlin2018} for BERT fine-tuning.

\subsection{Evaluation}

To evaluate the performance of the four systems, we calculate the span-level precision (P), recall (R) and F1 scores, along with corresponding micro and macro scores. 
The reported values are shown in Table~\ref{results} and are averaged over five folds, utilising the seqeval\footnote{\url{https://github.com/chakki-works/seqeval}} framework. 

With a macro avg.~F1-score of 0.59, MTL achieves the best result with a significant margin compared to CRF, BiLSTM-CRF and BERT. 
This confirms the usefulness of jointly training multiple objectives (minimizing multiple loss functions), and enabling knowledge transfer, especially in a setting with limited data (which is usually the case in the biomedical NLP domain). 
This result also suggest the usefulness of BioBERT for other biomedical datasets as reported by \newcite{Lee2019}. 
Despite being a rather standard approach, CRF outperforms the more elaborated BiLSTM-CRF, presumably due to data scarcity and class imbalance. 
We hypothesize that an increase in training data would yield better results for BiLSTM-CRF but not outperform transfer learning approach of MTL (or even BioBERT).
In contrast to other common NER corpora, like CoNLL 2003\footnote{\url{https://www.clips.uantwerpen.be/conll2003/ner/}}, even the best baseline system only achieves relatively low scores. 
This outcome is due to the inherent difficulty of the task (annotators are experienced medical doctors) and the small number of training samples.

\section{Conclusion}

We present a new corpus, developed to facilitate the processing of case reports. The corpus focuses on five distinct entity types: cases, conditions, factors, findings and modifiers. Where applicable, relationships between entities are also annotated. Additionally, we annotate discontinuous entities with a special relationship type (\textit{discontinuous}). 
The corpus presented in this paper is the very first of its kind and a valuable addition to the scarce number of corpora available in the field of biomedical NLP. Its complexity, given the discontinuous nature of entities and a high number of nested and multi-label entities, poses new challenges for NLP methods applied for NER and can, hence, be a valuable source for insights into what entities ``look like in the wild''. Moreover, it can serve as a playground for new modelling techniques such as the resolution of discontinuous entities as well as multi-task learning given the combination of entities and their relations. We provide an evaluation of four distinct NER systems that will serve as robust baselines for future work but which are, as of yet, unable to solve all the complex challenges this dataset holds.
A functional service based on the presented corpus is currently being integrated, as a NER service, in the QURATOR platform \cite{rehm2020d}. 

\section*{Acknowledgments} 
The research presented in this article is funded by the German Federal Ministry of Education and Research (BMBF) through the project QURATOR (Unternehmen Region, Wachstumskern, grant no.~03WKDA1A), see \url{http://qurator.ai}. 
We want to thank our medical experts for their help annotating the data set, especially Ashlee Finckh and Sophie Klopfenstein.

\section{Bibliographical References}
\label{main:ref}
\bibliographystyle{./lrec}
\bibliography{./main}

\end{document}